\pdfoutput=1
\documentclass[11pt]{article}
\PassOptionsToPackage{numbers,square,sort&compress}{natbib}
\usepackage[preprint]{acl}

\usepackage{times}
\usepackage{latexsym}
\usepackage[T1]{fontenc}
\usepackage[utf8]{inputenc}
\usepackage{microtype}
\usepackage{inconsolata}
\usepackage{graphicx}
\usepackage{booktabs}
\usepackage{placeins}  %
\usepackage{amsmath}
\usepackage{amssymb}
\usepackage{tikz}
\usetikzlibrary{positioning,arrows.meta,fit,backgrounds,calc,shapes.geometric}

\makeatletter
  {\bibitem@fin\bibpostamble\endlist}
\makeatother

\newcommand{\method}{DCKV}
\newcommand{\KL}{\mathrm{KL}}

\title{\method{}: Learned Transform Coding of the Prompt KV Cache\\in Hybrid Language Models}

\newif\iffullversion
\fullversionfalse

\author{Egor Maximov$^{1}$ \quad Aleksei Merkelov$^{1}$ \quad Vasiliy Alekseev$^{1}$ \quad Ivan Krivorotov$^{1}$ \\
  \textbf{Valentin Beliaev}$^{1}$ \quad \textbf{Nikita Arsenin}$^{2}$ \quad \textbf{Dmitry Bocharov}$^{2}$ \quad \textbf{Aleksei Goncharov}$^{1}$ \\
  $^{1}$MIL Team \quad $^{2}$DAIMLD}

\begin{document}
\maketitle

\begin{abstract}
Hybrid language models keep a key--value (KV) cache only in their few full-attention layers, but this cache is still the part of a prompt's state that grows with the prompt and that must be stored or moved to reuse it.
We present \method{}, a differentiable codec for the KV cache and the first KV-cache codec that is trained end to end.
Its transform, its quantizer and its entropy models, which are conditioned on the prompt tokens and on already decoded neighbouring positions, are optimized together against the frozen language model, both for the code length that the entropy models assign and for the change of the model's output.
On Qwen3.5-2B, \method{} stores a compressed token in more than 55 times fewer bytes than bf16 at a loss of at most 1.8 points on three long-context benchmarks, and in more than 100 times fewer bytes at a loss of at most 3.6 points.

\end{abstract}

\section{Introduction}
\label{sec:introduction}

Reusing the key--value (KV) cache of a long prompt saves its prefill, which dominates the time to the first token on long inputs.
To reuse prompt caches, serving systems keep them outside GPU memory and move them between devices and machines~\citep{liu2024cachegen,liu2025lmcache}.
Because such a cache is written once and read one or more times, a codec for it can afford entropy coding. What matters is the number of stored bytes at a given answer quality~\citep{staniszewski2026kvtc}.

Recent hybrid models interleave a few full-attention layers with linear-attention or state-space layers~\citep{lieber2024jamba,qwen2026qwen35}.
In Qwen3.5-2B, 6 of the 24 layers use full attention, and the other 18 are Gated DeltaNet layers~\citep{yang2025gdn} with a recurrent state of fixed size, 9.84~MiB per request.
Although the KV cache is small, 12,288 bytes per token in bf16, it is still the only part of the state that grows with the prompt, to 48~MiB at 4K tokens and 1.5~GiB at 128K.
We judge a codec by the memory of a stored request, recurrent state included.

KV cache transform coding (KVTC)~\citep{staniszewski2026kvtc} decorrelates keys and values with PCA, allocates bits by dynamic programming and applies a general-purpose lossless coder.
Its transform and its bit allocation minimize the reconstruction error of the cache and do not see the output of the model.
Learned image and video codecs train the transform and the quantizer on a rate--distortion objective~\citep{balle2017endtoend} and condition the entropy model on what the decoder already has~\citep{balle2018hyperprior,he2021checkerboard}.
For a language model, the natural distortion is the change of its output distribution, which some methods that compress the cache for decoding already use to learn cache projections~\citep{lin2025matryoshkakv}.

We present \method{}, a Differentiable Codec for the KV cache.
It is a learned codec for the stored prompt cache of a hybrid model and the first KV-cache codec whose transform, quantizer and entropy models are trained end to end.
Our contributions are:
\begin{itemize}
\item A codec with one joint transform for the keys and values of all attention layers, linear with a small nonlinear correction, with a learned quantization step for each component and GPU rANS coding, trained against the frozen model on the code length plus the KL divergence of the output (\S\ref{sec:method}).
\item Conditional entropy models that predict each code from the embedding of its token and from already decoded neighbouring tokens, trained together with the transform. With the correction they remove 12--16\% of the bytes at the same quality; fitted in closed form to a codec that is already trained, they give a similar saving on RULER 4K without retraining (\S\ref{sec:method-entropy}, \S\ref{sec:results-ablation}).
\item An evaluation on Qwen3.5-2B with paired per-task comparisons and a memory accounting that separates the state of a request from the codec weights (\S\ref{sec:setup}, \S\ref{sec:results}). Over three rates, \method{} needs 72\% fewer bytes than KVTC at equal score on RULER 4K and 62\% fewer on LongBench.
\end{itemize}

At prompts of 4K to 11K tokens the recurrent state takes about three quarters of the memory of a stored request. A smaller payload then reduces the request memory by a much smaller fraction than the stored cache.
At 32K tokens, where the payload of \method{} is more than a third of the request memory, \method{} scores 6.5 points higher than KVTC at about the same request memory (\S\ref{sec:results-matched}).

\section{Related Work}
\label{sec:related}

\paragraph{Hybrid models.}
Hybrid models replace most attention layers with layers whose state does not grow with the context, such as Mamba layers in Jamba~\citep{lieber2024jamba} or Gated DeltaNet layers~\citep{yang2025gdn} in Qwen3.5~\citep{qwen2026qwen35}.
Reusing a prompt in such a model needs both the attention cache and a snapshot of the recurrent state~\citep{pan2025marconi}.
The recurrent state has a fixed size and can be compressed by its own means, for example by quantization to a few bits~\citep{chen2026lowbit}, while the attention cache grows with the prompt and holds what the model must retrieve exactly~\citep{afendulev2026recall}.
\method{} compresses the attention cache and keeps the recurrent state exact, so methods for the recurrent state can be used together with it.

\paragraph{KV cache compression for decoding.}
Most methods shrink the cache while the model decodes, by evicting tokens~\citep{zhang2023h2o,li2024snapkv}, by quantizing keys and values~\citep{liu2024kivi,hooper2024kvquant}, by projecting them to a lower rank~\citep{chang2024palu,lin2025matryoshkakv}, or by sharing them across layers~\citep{brandon2024cla,chang2026xkv}.
Some of them learn their projection or their selection through the output of the model~\citep{lin2025matryoshkakv,roy2025kvcar}.
All of them keep a cache of fixed width that attention can read directly, and this requirement rules out codes of variable length~\citep{fernandez2026novakv}.
Their cache size is set by a design choice, such as a rank, a bit width or a number of tokens, and the statistical redundancy inside that width stays in the cache.
Because a stored cache is read only when its prompt is reused, an entropy coder can remove this redundancy, also after eviction or rank reduction.

\paragraph{Learned compression.}
Learned image codecs train a transform, a quantizer and an entropy model together on the rate plus $\lambda$ times the distortion~\citep{balle2017endtoend,balle2021ntc}.
Their entropy model is conditioned on information that the decoder already has, a side latent~\citep{balle2018hyperprior} or already decoded latents, which the checkerboard order decodes in parallel~\citep{he2021checkerboard}.
Video codecs add conditioning on the previous frame and serve several rates with one model~\citep{li2021dcvc,li2024dcvcfm}.
The same recipe has been used for intermediate features in split inference, where a bottleneck with a learned prior is trained into a vision model~\citep{matsubara2022supervised}.
For language models, an analysis of split transformer inference trains small models such as GPT-2 Small together with a codec for one hidden state~\citep{deandrade2026rd}.
In these works the network is trained together with its codec.
A codec for the KV cache of a deployed model must work with a model that is already trained, on an input that grows with the prompt.
\method{} is trained against the frozen model and works with a released model as it is. The same objective could also be added to pre-training.

\paragraph{Entropy-coded storage of KV caches.}
Codecs for a stored or transferred cache do use an entropy coder, and they differ in what comes before it.
Some quantize the cache directly or after a fixed rotation and code it with arithmetic, Huffman or rANS codes, under distributions that are profiled offline or fixed~\citep{liu2024cachegen,jiang2025kvcomp,su2026dquant}; KVCodec uses a hardware video codec~\citep{mi2026kvcodec}.
Others first remove the correlation in the cache, by PCA with bit allocation in KVTC~\citep{staniszewski2026kvtc}, by prediction from other layers or heads~\citep{shutova2025aquakv,yang2026alignpred}, or by bit allocation from rate--distortion theory~\citep{zuo2026ratequant}.
Codec-Gauge comes closest to a learned codec, as it trains channel transforms around an existing codec with a smooth proxy of the rate~\citep{jiang2026codecgauge}.
In all of them the probability model is fixed or profiled, and none trains its transform, quantizer and entropy model through the length of the code and the output of the model.
To our knowledge, \method{} is the first KV-cache codec that does, and its entropy models also use what the decoder of a stored prompt has, the prompt tokens and the already decoded neighbours.

\section{Method}
\label{sec:method}

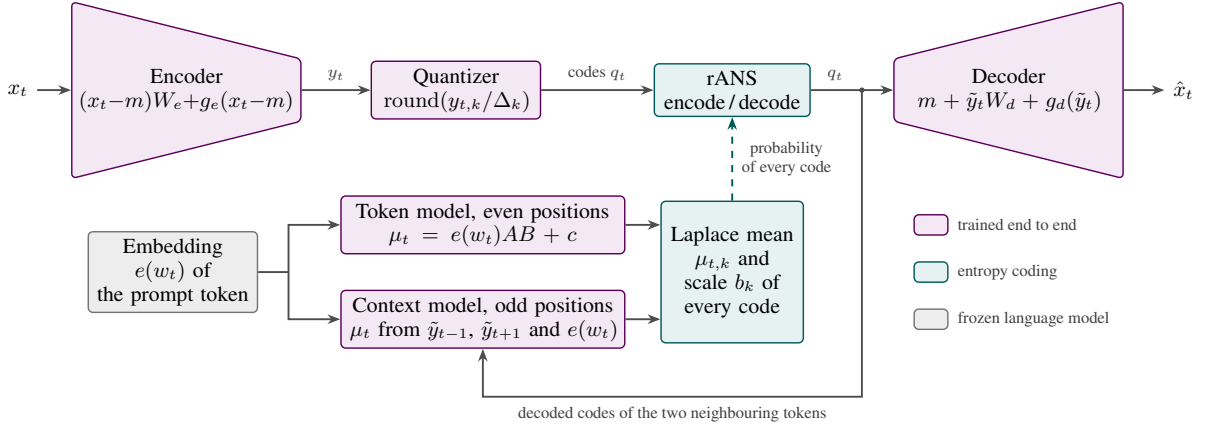
\begin{figure*}[t]
\centering
\resizebox{\textwidth}{!}{\begin{tikzpicture}[
    font=\footnotesize,
    >={Stealth[length=1.9mm]},
    trn/.style={draw=violet!70!black, fill=violet!9, rounded corners=2pt, align=center, inner sep=3pt, line width=0.6pt},
    ent/.style={draw=teal!70!black, fill=teal!10, rounded corners=2pt, align=center, inner sep=3pt, line width=0.6pt},
    frz/.style={draw=black!55, fill=black!7, rounded corners=2pt, align=center, inner sep=3pt, line width=0.6pt},
    flow/.style={->, line width=0.8pt, draw=black!70},
    lab/.style={font=\scriptsize, text=black!75, align=center},
]
\node (x) at (0,0) {$x_t$};
\node[trn, trapezium, trapezium angle=76, shape border rotate=270, inner xsep=0pt, inner ysep=4pt, text width=34mm] (enc) at (2.5,0)
    {Encoder\\$(x_t{-}m)W_e{+}g_e(x_t{-}m)$};
\node[trn, text width=23mm] (q) at (6.5,0) {Quantizer\\$\mathrm{round}(y_{t,k}/\Delta_k)$};
\node[ent, text width=21mm] (ans) at (10.6,0) {rANS\\encode\,/\,decode};
\node[trn, trapezium, trapezium angle=76, shape border rotate=90, inner xsep=0pt, inner ysep=4pt, text width=34mm] (dec) at (14.7,0)
    {Decoder\\$m + \tilde{y}_t W_d + g_d(\tilde{y}_t)$};
\node (xh) at (17.3,0) {$\hat{x}_t$};
\draw[flow] (x) -- (enc);
\draw[flow] (enc) -- node[lab, above] {$y_t$} (q);
\draw[flow] (q) -- node[lab, above] {codes $q_t$} (ans);
\draw[flow] (ans) -- node[lab, above, pos=0.3] {$q_t$} (dec);
\draw[flow] (dec) -- (xh);

\node[frz, text width=23mm] (emb) at (2.3,-2.7) {Embedding $e(w_t)$ of\\the prompt token};
\node[trn, text width=40mm] (tok) at (6.9,-2.0) {Token model, even positions\\$\mu_t = e(w_t) A B + c$};
\node[trn, text width=40mm] (ctx) at (6.9,-3.4) {Context model, odd positions\\$\mu_t$ from $\tilde{y}_{t-1}$, $\tilde{y}_{t+1}$ and $e(w_t)$};
\node[ent, text width=19mm, minimum height=21mm] (par) at (10.6,-2.7) {Laplace mean\\$\mu_{t,k}$ and\\scale $b_k$ of\\every code};
\coordinate (fork) at ($(emb.east)+(0.45,0)$);
\draw[line width=0.8pt, draw=black!70] (emb.east) -- (fork);
\draw[flow] (fork) |- (tok.west);
\draw[flow] (fork) |- (ctx.west);
\draw[flow] (tok.east) -- (tok.east -| par.west);
\draw[flow] (ctx.east) -- (ctx.east -| par.west);
\draw[flow, dashed, draw=teal!70!black] (par.north) -- node[lab, right] {probability\\of every code} (ans.south);
\coordinate (tap) at ($(ans.east)!0.62!(dec.west)$);
\fill[black!70] (tap) circle (1.1pt);
\draw[flow] (tap) -- (tap |- 0,-4.55) -- node[lab, below] {decoded codes of the two neighbouring tokens} (ctx.south |- 0,-4.55) -- (ctx.south);

\node[trn, minimum width=5mm, minimum height=3mm, inner sep=0pt] (l1) at (13.55,-2.0) {};
\node[lab, anchor=west] at (l1.east) {trained end to end};
\node[ent, minimum width=5mm, minimum height=3mm, inner sep=0pt] (l2) at (13.55,-2.7) {};
\node[lab, anchor=west] at (l2.east) {entropy coding};
\node[frz, minimum width=5mm, minimum height=3mm, inner sep=0pt] (l3) at (13.55,-3.4) {};
\node[lab, anchor=west] at (l3.east) {frozen language model};
\end{tikzpicture}}
\caption{\method{}. Top: the cache row $x_t$ of a token is transformed, quantized and entropy-coded once after prefill, and restored once before decoding from its dequantized coefficients $\tilde{y}_t = q_t \odot \Delta$. Bottom: the conditional models tell the entropy coder the mean and the scale of every code from what the decoder already has: the prompt token and, at odd positions, the decoded codes of the two neighbouring tokens.}
\label{fig:overview}
\end{figure*}

\subsection{What is compressed}
\label{sec:method-setting}

\method{} compresses the attention cache of a prefilled prompt once and restores it completely before decoding (Figure~\ref{fig:overview}); the recurrent state is kept exact.
As in KVTC, the first 4 and the last 128 cache positions stay in bf16, and the other positions form the compressed region $\mathcal{T}$.
For a token $t \in \mathcal{T}$ we stack the keys and values of all $L=6$ attention layers into one row $x_t \in \mathbb{R}^{d}$, with $d = 2 L H d_h = 6144$ for $H=2$ key--value heads of dimension $d_h=256$.
To make the row independent of the absolute position, we take the keys before the rotary position embedding (RoPE) and apply it again after decompression.

\subsection{Transform and quantization}
\label{sec:method-transform}

An encoder maps a row to $n=2048$ coefficients, and a decoder maps the quantized coefficients back:
\begin{align}
y_t &= (x_t - m)\,W_e + g_e(x_t - m), \nonumber\\
q_{t,k} &= \mathrm{round}\!\left(y_{t,k}/\Delta_k\right), \qquad \tilde{y}_t = q_t \odot \Delta, \nonumber\\
\hat{x}_t &= m + \tilde{y}_t\,W_d + g_d(\tilde{y}_t), \label{eq:codec}
\end{align}
where $m$ is the mean row of the calibration data, $W_e \in \mathbb{R}^{d\times n}$ and $W_d \in \mathbb{R}^{n\times d}$ are matrices, and $\Delta \in \mathbb{R}^{n}_{>0}$ holds one quantization step per component.
The matrices start from the leading $n$ principal directions $V$ of calibration rows, $W_e = V$ and $W_d = V^{\top}$, with equal steps. Before training, the codec is PCA with a uniform quantizer, which we call the untrained joint basis.
Unlike KVTC, which keeps one basis for the keys and one for the values, \method{} uses a single basis for both and for all layers.
The nonlinear correction $g_e$ is a network with one hidden layer of 256 units and a GELU, and $g_d$ mirrors it on the decoder side; the last layer of each starts at zero, so training starts from the linear transform.
We call a codec without the correction and with the static entropy model of \S\ref{sec:method-coding} a linear codec. The linear codec from an earlier training run is the main ablation.
A component whose scale in units of its step, $\beta_k$ in Eq.~\eqref{eq:laplace}, is below 0.05 is switched off and is neither stored nor reconstructed.

\subsection{Entropy coding}
\label{sec:method-coding}

Each code $q_{t,k}$ is written with rANS~\citep{duda2013ans,giesen2014interleaved}.
The static entropy model is a zero-mean Laplace distribution per component, integrated over the quantization cell:
\begin{equation}
P(q_{t,k}) = F_{\beta_k}\!\left(q_{t,k} + \tfrac{1}{2}\right) - F_{\beta_k}\!\left(q_{t,k} - \tfrac{1}{2}\right),
\label{eq:laplace}
\end{equation}
where $F_\beta$ is the Laplace distribution function with scale $\beta_k = b_k/\Delta_k$, and $b_k$ starts at the mean absolute coefficient on calibration rows.
Codes beyond $\pm 127$ are escaped and stored separately, at 64 bits each.
We use the GPU rANS of the \texttt{torch-ans} library\footnote{\url{https://github.com/worldlife123/torch_ans}} with 15-bit frequency tables and interleaved streams of 16,384 symbols.
On held-out text the stored payload is 5\% (linear codec) and 17\% (\method{}) above the code length that the trained entropy models expect (Appendix~\ref{app:details}).

\subsection{Conditional entropy models}
\label{sec:method-entropy}

The static model ignores what the decoder already knows.
\method{} replaces it with two conditional models, which shift the Laplace distribution of each code by a predicted mean.

\paragraph{Token-conditioned model.}
The decoder knows the prompt tokens, because a request that reuses a stored prompt cache contains the prompt text. We do not count the token ids in the payload (as 4-byte integers they would add 1--5\% at our rates).
The model predicts the coefficients of token $t$ from the input embedding $e(w_t)$ of its token, $\mu_t = e(w_t)\,A B + c$, with $A \in \mathbb{R}^{2048\times r}$ and $B \in \mathbb{R}^{r\times n}$. \method{} uses the full $r = 2048$, although a smaller $r$ would save weights. The residual has a Laplace distribution with one trained scale per component.
The code is written as the symbol $q_{t,k} - \mathrm{round}(\mu_{t,k}/\Delta_k)$ under one of 576 precomputed tables (0.84~MiB), one for each pair of 64 levels of the residual scale and 9 offsets of the fractional part of the predicted mean.
Encoder and decoder select the same table as long as they compute the same predictions; all our evaluations encode and decode in the same process (see Limitations).

\paragraph{Checkerboard context model.}
The codes of neighbouring tokens are correlated, but a model that predicts every token from the one before it has to decode the tokens one by one.
We instead split the positions of the compressed region into even and odd ones, the one-dimensional form of the checkerboard order of image codecs~\citep{he2021checkerboard}.
Even positions are coded first, with the token-conditioned model alone.
Odd positions are coded second, when both of their neighbours are already decoded. A linear map of rank 1024 predicts the mean of the codes of an odd position from the dequantized coefficients of its two neighbours and from the embedding of its own token.
Each pass is parallel over tokens, so a cache is decoded in two parallel steps for any length.

\subsection{Rate--distortion training}
\label{sec:method-training}

We train $W_e$, $W_d$, $g_e$, $g_d$, $\Delta$ and the entropy models against the frozen language model; $m$ stays fixed.
A training sequence has a prompt of $P=4096$ tokens, prefilled once without gradients, and a continuation of $C=256$ tokens.
The compressed region of the prompt passes through the codec, and the model reads the continuation twice, once over the exact cache, which gives the reference next-token distributions $p_i$, and once over the reconstructed cache, which gives $\hat{p}_i$.
The loss is
\begin{equation}
\mathcal{L} = \frac{R}{|\mathcal{T}|} + \lambda\,\frac{1}{C}\sum_{i=1}^{C} \KL\!\left(p_i \,\|\, \hat{p}_i\right),
\label{eq:loss}
\end{equation}
where $R$ is the number of bits of the region under the conditional models (under Eq.~\eqref{eq:laplace} for the linear codec) and the KL divergence is in nats.
Because the rate is measured under the conditional models, the transform learns to leave to the codes what these models cannot predict.
The conditional models start from a closed-form fit to the untrained codec, and several codecs train in one loop on shared prefills.

Rounding has a zero gradient almost everywhere.
The usual proxy adds uniform noise $u \sim \mathcal{U}(-\tfrac12,\tfrac12)$ to $y_{t,k}/\Delta_k$ in both terms of the loss~\citep{balle2017endtoend}.
For a component with $\beta_k \le 0.2$, rounding sets more than 90\% of the coefficients to zero. Because the noisy coefficients still carry their signal to the decoder during training, we expect the KL term to underestimate the distortion of the real codec. The noise-based rate, in turn, overestimates the entropy of the rounded code by a factor of 1.47 at $\beta_k = 0.2$ and 5.66 at $\beta_k = 0.1$ (Appendix~\ref{app:rate-bias}).
We therefore use mixed quantization~\citep{minnen2020channel}, which computes the rate on the noisy coefficient and reconstructs from the rounded coefficient with a straight-through gradient~\citep{bengio2013estimating,theis2017lossy}.
This estimator can degrade nonlinear transforms at low rates~\citep{balle2021ntc}; our correction is trained with it and gains most at low rates (Appendix~\ref{app:offline}).

A codec trained with one $\lambda$ serves one rate.
Two cheaper ways to change the rate, a factor on all steps of a trained codec and one codec trained for a range of $\lambda$, cost bytes or quality (Appendix~\ref{app:offline}).

\subsection{Memory accounting}
\label{sec:method-memory}

The request memory is what a stored request needs besides the model: the payload, the 132 uncompressed positions (1.55~MiB) and the recurrent state (9.84~MiB); for the uncompressed model, the full attention cache and the same recurrent state.
Since codec weights are loaded once with the model and shared by all requests, we count them as model weights for all codecs. We report them separately: 176~MiB of fp32 parameters for \method{}, about 5\% of the weights of the 2B model, 96~MiB for the linear codec and 10--20~MiB for KVTC.
Peak GPU memory grows by 182~MiB with \method{} and by 102~MiB with the linear codec (Appendix~\ref{app:efficiency}).

\section{Experimental Setup}
\label{sec:setup}

\paragraph{Model and benchmarks.}
We use Qwen3.5-2B~\citep{qwen2026qwen35} in bf16 with greedy decoding, one request at a time, with Hugging Face Transformers on one A100 40GB GPU; generative benchmarks use the chat template in non-thinking mode.
RULER~\citep{hsieh2024ruler} at 4K tokens has 13 tasks, 1,287 samples and a mean prompt of 3,742 tokens, and we score it by token recall, the fraction of the words of the reference answer found in the generated answer.
RULER at 32K tokens has the same 13 tasks, 1,297 samples and its own uncompressed reference.
From LongBench~\citep{bai2024longbench} we use 18 of the 21 tasks, with a mean prompt of 10.8K tokens and a re-implementation of the official metrics.
We leave out LCC and RepoBench-P, on which no run of our codecs, of KVTC or of the uncompressed model scores more than 4.4 points, and SAMSum, on which, as on RepoBench-P, every run of our codecs and of KVTC scores above the uncompressed reference (31 and 30 runs; Appendix~\ref{app:per-task}).
If we include all 21 tasks (20 for \method{} at $\lambda = 5{\cdot}10^4$), every loss is 0.3--1.0 points smaller and the order of the methods does not change.
MMLU~\citep{hendrycks2021mmlu} has 57 subjects, 5-shot, scored by the likelihood of the answer options, with a mean prompt of 620 tokens.
Appendix~\ref{app:short} gives two benchmarks with shorter prompts, GPQA Diamond and IFEval.

\paragraph{Baselines.}
We found no official implementation of KVTC~\citep{staniszewski2026kvtc}.
We start from a public proof of concept\footnote{\url{https://github.com/bet0x/kvtc-poc}}, which allocates bits greedily, and replace this allocation with the dynamic programming of the paper.
At $16\times$ the greedy allocation loses 13.0 points on RULER 4K, and the dynamic programming loses 0.53.
Our KVTC uses two PCA bases of rank 1024, for the keys before RoPE and for the values of all attention layers, bit allocation for a nominal compression ratio, and DEFLATE.
Its ratios of $16\times$, $24\times$ and $32\times$ give 627, 410 and 304 bytes per token on RULER 4K.
We allow all widths from 0 to 8 bits. Restricting the code widths to 0, 2, 4 and 8 bits, the widths closest to the data types of the KVTC paper, saves 6--10\% of the bytes at $8\times$ and $16\times$ with a slightly smaller loss, but at $32\times$ it loses 7.2 points on RULER 4K instead of 3.9.
xKV~\citep{chang2026xkv}, a cache compression method for decoding, projects the keys and values of each pair of adjacent attention layers onto a shared low-rank basis from a singular value decomposition; we use rank 192 for keys and 288 for values.
xKV computes the basis for each prompt, so we count the basis (2.8~MiB) in the request memory and in the payload, together with the 16-bit coefficients of all prompt tokens.
The ablations use the trained \emph{linear codec} (\S\ref{sec:method-transform}) and two untrained codecs with the quantizer and coder of \method{}. One keeps KVTC's separate key and value bases with one uniform step (\emph{separate bases}), and the other is the \emph{untrained joint basis}.

\paragraph{Training.}
The codecs of Table~\ref{tab:main} start from a joint PCA basis with all steps at 1.0. The basis comes from up to 200K token positions of 48 calibration documents of 1K to 16K tokens each, 32 of web text (FineWeb) and 16 of mathematical reasoning (OpenR1-Math). The same documents calibrate KVTC.
They train on FineWeb-Edu~\citep{penedo2024fineweb} with AdamW~\citep{loshchilov2019adamw} on two A100 GPUs for 1B tokens.
\method{} at $\lambda = 3{\cdot}10^5$ trains in one loop with its three ablations (\S\ref{sec:results-ablation}), which took 33.9 hours.
\method{} at $\lambda = 10^5$ and $5{\cdot}10^4$ and the codec without additions (no correction and no conditional models) at $10^5$ train in a second loop with the same recipe, which took 25.9 hours.
The linear codecs of Table~\ref{tab:main} come from an earlier run of three codecs, which took about 17 hours (Appendix~\ref{app:details}).
A run with the noise proxy used $\lambda \in \{3{\cdot}10^4, 10^5, 3{\cdot}10^5\}$, and we report its checkpoints at 0.7B tokens.

\paragraph{Comparisons.}
A score change is the paired difference of benchmark means over the tasks that both runs share, reported with the number of tasks $n$.
For two codecs we also count the tasks on which the first is better, equal and worse (W/T/L) and apply a two-sided sign test over the tasks that differ; Table~\ref{tab:pairs} in Appendix~\ref{app:pairs} lists these statistics.
The evaluation is deterministic, and repeated launches of a configuration, also on two versions of the evaluation code and on two GPUs in parallel, give identical scores on every task.
The only uncertainty in a difference between two runs comes from the finite numbers of samples and tasks; two uncompressed runs with about 50 and 100 samples per task differ by 0.09 points on the RULER 4K mean and by up to 4 points on single tasks.
For two rate--quality curves we also give the Bjøntegaard delta rate (BD-rate)~\citep{bjontegaard2001}, the mean difference of the log payload at equal score over the range of scores that both curves cover, in percent.
With three to five points per curve, we interpolate linearly and read the value as an estimate~\citep{barman2024bd}.
Bytes per token count the compressed region only.

\section{Results}
\label{sec:results}

\begin{table*}[t]
\centering
\footnotesize
\setlength{\tabcolsep}{3.0pt}
\begin{tabular}{l rrr rrrr}
\toprule
 & \multicolumn{3}{c}{Payload, bytes per token} & \multicolumn{4}{c}{Score change vs.\ uncompressed, points} \\
\cmidrule(lr){2-4}\cmidrule(lr){5-8}
Method & RULER 4K & LongBench & RULER 32K & RULER 4K & LongBench & MMLU & RULER 32K \\
 & & & & $n{=}13$ & $n{=}18$ & $n{=}57$ & $n{=}13$ \\
\midrule
Uncompressed & 12,288 & 12,288 & 12,288 & 94.54 & 42.66 & 59.34 & 94.25 \\
\midrule
KVTC, $16\times$ & 627 & 633 & 548 & \textbf{-0.53} & \textbf{-0.31} & -0.62 & \textbf{-0.02} \\
KVTC, $24\times$ & 410 & 415 & 356 & -1.87 & -1.97 & -0.99 & -2.45 \\
KVTC, $32\times$ & 304 & 307 & 263 & -3.88 & -4.17 & -0.93 & -6.94 \\
xKV, ranks 192/288 & 3,667 & 3,151 & -- & -2.65 & -3.73 & -0.42 & -- \\
\midrule
DCKV, $\lambda{=}3{\cdot}10^{5}$ & 214 & 205 & 221 & \textbf{-0.56} & \textbf{-1.75} & -1.10 & \textbf{-0.45} \\
DCKV, $\lambda{=}10^{5}$ & 113 & 114 & 122 & \textbf{-1.69} & \textbf{-3.55} & -0.87 & \textbf{-2.15} \\
DCKV, $\lambda{=}5{\cdot}10^{4}$ & 74 & 80 & 83 & \textbf{-4.60} & \textbf{-5.51} & -0.65 & \textbf{-6.45} \\
\midrule
Linear codec, $\lambda{=}3{\cdot}10^{5}$ & 282 & 273 & 288 & -0.60 & \textbf{-1.25} & -0.79 & \textbf{-0.08} \\
Linear codec, $\lambda{=}10^{5}$ & 159 & 154 & 163 & -2.28 & \textbf{-2.85} & -0.83 & -2.29 \\
Linear codec, $\lambda{=}5{\cdot}10^{4}$ & 108 & 105 & 111 & -5.74 & \textbf{-4.67} & -0.78 & \textbf{-6.14} \\
\bottomrule
\end{tabular}

\caption{Payload (bytes per compressed token) and score change against the uncompressed model. The first row gives absolute scores; $n$ is the number of tasks. Every \method{} and linear-codec row is a codec trained for its $\lambda$. Linear codec: no nonlinear correction, static entropy model. Bold: on each long-context benchmark, the runs that no other compressed run beats in both payload and score. Table~\ref{tab:efficiency} in Appendix~\ref{app:efficiency} gives their request memory, and Appendix~\ref{app:short} the benchmarks with shorter prompts. --: not run.}
\label{tab:main}
\end{table*}

\subsection{Quality and memory}
\label{sec:results-main}

Table~\ref{tab:main} gives the payload and the score change of every run, and Figure~\ref{fig:rate-quality} plots one against the other.
Table~\ref{tab:pairs} in Appendix~\ref{app:pairs} gives the paired statistics of the comparisons in this section.

At $\lambda = 3{\cdot}10^5$, \method{} stores 205 to 221 bytes per token, 56 to 60 times fewer bytes than bf16, and loses 0.56, 1.75 and 0.45 points on RULER 4K, LongBench and RULER 32K.
KVTC at $16\times$ loses at most 0.53 points but stores 2.5 to 3.1 times as many bytes, and on LongBench it is 1.44 points better than \method{} ($p{=}0.03$).

The advantage of \method{} grows at stronger compression.
At $\lambda = 10^5$ it stores 113 to 122 bytes per token and scores higher than KVTC at $32\times$ on all three benchmarks, although KVTC stores 2.2 to 2.7 times as many bytes; the difference is significant on LongBench and RULER 32K.
At $\lambda = 5{\cdot}10^4$, \method{} does not differ significantly from KVTC at $32\times$ on any of the three benchmarks and needs about a quarter to a third of its bytes.
Over the three rates, \method{} needs 72\% fewer bytes than KVTC at equal score on RULER 4K and 62\% fewer on LongBench (BD-rate).
The trained linear codec, without the correction and the conditional models, already needs less than half the bytes of KVTC at equal score (BD-rate $-56\%$ and $-55\%$).

xKV~\citep{chang2026xkv} stores 3,667 and 3,151 bytes per token, its basis for the prompt included, and loses 2.65 and 3.73 points on RULER 4K and LongBench.
\method{} at $\lambda = 10^5$ does not differ significantly from it and needs 96--97\% fewer bytes.
MMLU has a mean prompt of 620 tokens, and on it every run of \method{}, the linear codec and KVTC loses 0.5 to 1.5 points, whatever its payload, so this benchmark hardly separates the codecs.

\begin{figure*}[t]
\centering
\includegraphics[width=\textwidth]{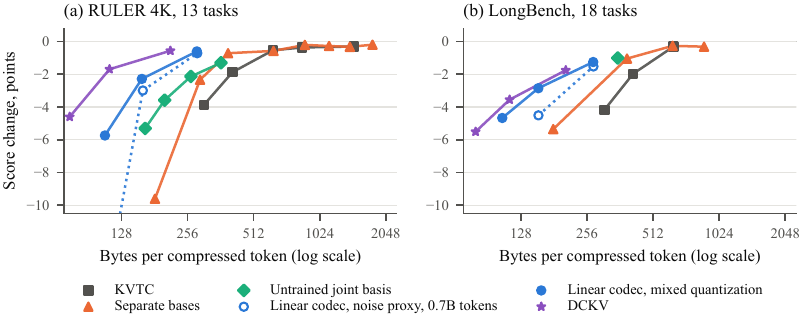}
\caption{Score change against the payload (bytes per compressed token; bf16 takes 12,288). Lines connect the runs of one family. Not shown: xKV (3,151 and 3,667 bytes per token) and, in (a), the linear codec with the noise proxy at $\lambda{=}3{\cdot}10^4$ (79 bytes, $-25.3$ points).}
\label{fig:rate-quality}
\end{figure*}

\begin{table}[t]
\centering
\footnotesize
\setlength{\tabcolsep}{2.7pt}
\begin{tabular}{l rr rr}
\toprule
 & \multicolumn{2}{c}{RULER 4K} & \multicolumn{2}{c}{LongBench} \\
\cmidrule(lr){2-3}\cmidrule(lr){4-5}
Configuration & B/tok. & Change & B/tok. & Change \\
\midrule
KVTC $24\times$ & 410 & -1.87 & 415 & -1.97 \\
Separate bases, step 2.0 & 392 & -0.71 & 389 & -1.04 \\
Joint basis, step 2.0 & 364 & -1.30 & 353 & \textbf{-1.00} \\
Joint basis, step 2.7 & \textbf{265} & -2.13 & -- & -- \\
Trained, noise, $\lambda{=}3{\cdot}10^{5}$ & 283 & -0.71 & 274 & -1.52 \\
Trained, mixed, $\lambda{=}3{\cdot}10^{5}$ & 282 & \textbf{-0.60} & \textbf{273} & -1.25 \\
\midrule
No additions & 245 & -0.96 & 237 & \textbf{-1.73} \\
\quad + nonlinear correction & 232 & -0.71 & -- & -- \\
\quad + trained cond.\ models & 224 & -0.59 & -- & -- \\
\quad + both: DCKV & \textbf{214} & \textbf{-0.56} & \textbf{205} & -1.75 \\
\midrule
Separate bases, step 4.0 & 182 & -9.61 & 180 & -5.35 \\
Joint basis, step 4.0 & 164 & -5.31 & -- & -- \\
Trained, noise, $\lambda{=}10^{5}$ & 160 & -2.99 & \textbf{154} & -4.51 \\
Trained, mixed, $\lambda{=}10^{5}$ & \textbf{159} & \textbf{-2.28} & \textbf{154} & \textbf{-2.85} \\
\bottomrule
\end{tabular}

\caption{From KVTC to \method{}: bytes per token and score change. First block: 260--420 bytes per token; second: one run that trains four codecs at $\lambda{=}3{\cdot}10^5$; third: 150--185 bytes. Bold: the fewest bytes and the best score change of each benchmark within a block.}
\label{tab:ablation}
\end{table}

\subsection{Where the gains come from}
\label{sec:results-ablation}

Table~\ref{tab:ablation} adds the parts of \method{} one at a time.

\paragraph{Entropy coding instead of bit allocation.}
We keep KVTC's separate key and value bases and replace its dynamic-programming allocation and DEFLATE with one quantization step, a Laplace model and rANS.
This gains about one point on RULER 4K and LongBench at a slightly smaller payload.

\paragraph{Joint basis.}
One basis for keys and values needs 7--10\% fewer bytes than separate bases at the same step.
At step 2.0 the scores differ by less than 0.6 points, and at the coarse step 4.0 the joint basis scores 4.3 points higher on RULER 4K, although this difference is not significant.

\paragraph{Rate--distortion training.}
At a similar payload, the trained linear codec at $\lambda = 10^5$ scores 3.0 points higher on RULER 4K than the untrained joint basis (159 against 164 bytes per token).
Mixed quantization matters at stronger compression.
At $\lambda = 10^5$ it scores 0.7 and 1.7 points higher than the noise proxy on RULER 4K and LongBench, and it is better on 20 of the 31 tasks and worse on 6.
At $\lambda = 3{\cdot}10^5$ the two estimators do not differ significantly.

\paragraph{Correction and trained conditional models.}
One training run of 1B tokens at $\lambda = 3{\cdot}10^5$ trains four codecs in one loop from the untrained joint basis.
They are a codec without additions, one with the correction, one with conditional models, and \method{} with both.
The codec without additions has the form of the linear codec of Table~\ref{tab:main} but comes from this run (245 against 282 bytes per token on RULER 4K).
\method{} stores 12--14\% fewer bytes than the codec without additions on the three benchmarks, and the scores do not differ significantly.
On RULER 4K, the correction alone saves 5.5\% of the bytes and the conditional models alone 8.7\%.
In a second run at $\lambda = 10^5$, the additions save 16\% and 13\% of the bytes on RULER 4K and LongBench, and the scores are 1.1 and 0.8 points higher (not significant).

Because rounding does not depend on the entropy model, conditional models can also be fitted in closed form to a codec that is already trained, and no answer changes.
Fitted to the codec without additions at $\lambda = 3{\cdot}10^5$, they save 12.5\% of the bytes on RULER 4K, as much as the trained additions.
Training the additions together with the transform gains more at stronger compression. Among codecs trained for a range of $\lambda$, \method{} needs 22\% fewer bytes at equal score than the codec without additions with fitted models (Appendix~\ref{app:offline}).

\begin{figure}[t]
\centering
\includegraphics[width=\columnwidth]{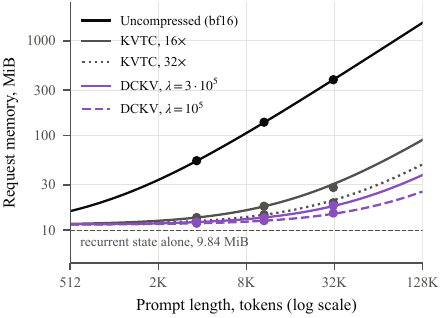}
\caption{Request memory against prompt length, computed from the payload per token on RULER 4K. Markers show measured values on RULER 4K, LongBench and RULER 32K (mean prompts of 3.7K, 10.8K and 32.2K tokens).}
\label{fig:request-memory}
\end{figure}

\subsection{Similar request memory}
\label{sec:results-matched}

The recurrent state takes 81\% of the request memory of \method{} on RULER 4K, 73\% on LongBench and 54\% on RULER 32K (Figure~\ref{fig:request-memory}; Tables~\ref{tab:matched} and~\ref{tab:efficiency} in Appendix~\ref{app:efficiency}).
On RULER 4K, KVTC needs at most 12\% more request memory than \method{}, and no difference between them is significant.
On LongBench and RULER 32K, \method{} scores 2.4 and 6.5 points higher than KVTC at $32\times$, which needs 8\% and 7\% more request memory.
Only KVTC at $16\times$ is not significantly worse than \method{} there (on LongBench it is 1.44 points better), and it needs 32\% and 55\% more request memory.
At $\lambda = 10^5$, \method{} needs 5--22\% less request memory than KVTC at $32\times$ and still scores higher on all three benchmarks.

\subsection{Behaviour by task}
\label{sec:results-tasks}

Figure~\ref{fig:per-task} in Appendix~\ref{app:per-task} shows the change of every task.
On RULER 4K one question-answering task loses 3.3 points with \method{} and no other task more than 1.3.
On RULER 32K, variable tracking loses 4.2 points and the same question-answering task 3.7.
On LongBench, passage retrieval loses most, 10.0 points in English and 6.0 in Chinese, against 4.0 and 1.0 with KVTC at $16\times$ and 11.5 and 12.5 at $32\times$.
At $\lambda = 10^5$, variable tracking loses most on RULER, 5.8 points at 4K and 11.0 at 32K.
Common-word extraction loses 3.4 points at 4K, and a needle with a UUID value loses 3.0 at 4K and 6.0 at 32K.
On LongBench, passage retrieval loses 27.0 points in English and 12.5 in Chinese.

\section{Conclusion}
\label{sec:conclusion}

\method{} trains the transform, the quantizer and the conditional entropy models of a codec for the stored prompt cache of a hybrid language model end to end, against the frozen model.
On Qwen3.5-2B it stores the cache in about 210 bytes per token, 2.9 times fewer than KVTC at the same loss on RULER 4K, and its advantage grows at stronger compression.
Over three rates it needs about a third of the bytes of KVTC at equal score on RULER 4K and LongBench (\S\ref{sec:results-main}).
The ablations show that each part adds to this gain, from entropy coding and the joint basis to the training against the output of the model and the conditional entropy models (\S\ref{sec:results-ablation}).

A smaller payload matters most for long prompts, because at moderate lengths the recurrent state takes most of the memory of a stored request; at 32K tokens \method{} scores higher than KVTC at about the same request memory (\S\ref{sec:results-matched}).
Since the codec is trained against a frozen model, it can serve a released model without changes to its weights.
Open questions are whether the results carry over to other hybrid models, how to train one codec for a range of rates without a loss at the upper end, and how to make decoding give identical results on different devices.
A model that is pre-trained together with the codec, with the code length of its cache in the loss, may give a cache that compresses even better.

\section*{Limitations}

\paragraph{One model, few tasks.}
We developed the codec on Qwen3.5-0.8B, but all results are for Qwen3.5-2B, and we did not test other hybrid families.
The evaluation is deterministic, but with 13 and 18 tasks many differences between strong codecs are not significant under the sign test, and we did not compute intervals over samples or correct the p-values for multiple comparisons.
We cannot yet say why a few tasks, such as variable tracking and passage retrieval, lose much more than the others, because we have not analysed the answers sample by sample.

\paragraph{Rate control.}
Each rate needs its own codec and its own training run.
We also implemented and tested one codec for a range of $\lambda$ with the mechanism of variable-rate video codecs~\citep{li2024dcvcfm}.
It falls behind the codecs for one $\lambda$ at the upper end of the range, and we did not find the cause (Appendix~\ref{app:offline}).
A factor on the steps of a trained codec changes the rate without training but costs bytes.

\paragraph{Frozen model.}
We train the codec for a frozen language model, which lets it work with a released model as it is.
Continued pre-training can make a KV cache easier to compress for post-hoc methods~\citep{gelberg2026compressibility}. A model that is pre-trained together with the codec, with the code length of its cache in the loss, may also give a cache that compresses better, but we did not have the compute to test it.

\paragraph{Context length.}
The codecs were trained on 4K-token prompts and evaluated on prompts of up to 32K tokens.
Beyond 32K tokens the curves of Figure~\ref{fig:request-memory} are computed, not measured.
At 32K tokens the measured payload per token is 3\% above that on RULER 4K for \method{} and 12--14\% below it for KVTC, so the computed curves overstate the advantage of \method{}.

\paragraph{Codec weights and peak memory.}
Because \method{} restores the full bf16 cache before decoding, it does not reduce peak GPU memory or the time per output token.
Its weights add about 5\% to the size of the 2B model, and peak GPU memory grows by about the size of these weights (\S\ref{sec:method-memory}).

\paragraph{Latency.}
We timed the uncompressed model, KVTC at $16\times$ and \method{} once, one after another on one A100 of a shared node (Table~\ref{tab:timing}, Appendix~\ref{app:efficiency}).
With a stored cache, the time to the first token, which includes decompression, is 115~ms for \method{} at 4K tokens and 287~ms at 32K, against 862 and 4,601~ms with the prefill of the uncompressed model.
We did not time the compression alone, and the time per output token does not change.
The time of KVTC on our shared node is not stable, with 6.8~s at 32K in this run and 0.6~s in an earlier one.

\paragraph{Baselines and evaluation.}
Our KVTC has one quantization range per component for the whole prompt and no block sizes or data types; the paper chooses them per block of components, and we did not evaluate that variant or compare with the numbers of the paper.
On MMLU every run of \method{}, the linear codec and KVTC loses 0.5--1.5 points whatever its payload. This loss may come from the part of our evaluation harness that runs a compressed cache rather than from the compression.
On IFEval and on two LongBench tasks that we leave out of the means, every run of our codecs and of KVTC scores above the uncompressed reference (Appendices~\ref{app:short} and~\ref{app:per-task}).
We did not find the cause of these offsets.

\paragraph{Decoding on another device.}
All our evaluations encode and decode in the same process.
Since the conditional models compute their predictions in floating point, decoding on another device needs predictions that are identical bit for bit, for example from integer inference as in DCVC-RT~\citep{jia2025dcvcrt}; we have not verified this.

\section*{Acknowledgments}
We used an AI assistant for editing the text and for checking the reported numbers against the experiment logs.

\bibliography{references}

\appendix
\onecolumn  %
\section{Bias of the Uniform-Noise Rate Estimate}
\label{app:rate-bias}

Let a coefficient divided by its step follow $Y \sim \mathrm{Laplace}(0, \beta)$, and let $U \sim \mathcal{U}(-\tfrac12, \tfrac12)$ be independent of it.
The density of $Z = Y + U$ is $g(z) = F_\beta(z + \tfrac12) - F_\beta(z - \tfrac12)$, the probability that Eq.~\eqref{eq:laplace} assigns to the cell around $z$.
The rate estimated during training is therefore $\mathbb{E}[-\log_2 g(Z)] = h(Z)$, the differential entropy of $Z$.
The rate of the rounded code $Q = \mathrm{round}(Y)$ is its entropy $H(Q) = -\sum_j P_j \log_2 P_j$ with $P_j = F_\beta(j + \tfrac12) - F_\beta(j - \tfrac12)$.
A code differs from zero with probability $P(|Y| > \tfrac12) = e^{-1/(2\beta)}$, so for small $\beta$ the entropy $H(Q)$ vanishes exponentially, while $h(Z)$ decreases only about linearly in $\beta$.
Table~\ref{tab:rate-bias} gives both quantities, computed by numerical integration.

\begin{table}[!ht]
\centering
\small
\setlength{\tabcolsep}{4pt}
\begin{tabular}{rrrrr}
\toprule
$\beta$ & $P(Q \neq 0)$ & $H(Q)$ & $h(Z)$ & $h(Z)/H(Q)$ \\
\midrule
1.0 & 0.607 & 2.484 & 2.492 & 1.00 \\
0.5 & 0.368 & 1.560 & 1.606 & 1.03 \\
0.2 & 0.082 & 0.496 & 0.730 & 1.47 \\
0.1 & 0.007 & 0.065 & 0.368 & 5.66 \\
\bottomrule
\end{tabular}
\caption{Entropy of the rounded code, $H(Q)$, and its uniform-noise estimate, $h(Z)$, in bits, for a Laplace coefficient with scale $\beta$ in units of the step.}
\label{tab:rate-bias}
\end{table}

In the training run of the linear codecs, the rate estimated during training exceeds the rate measured on held-out data with real rounding by 45\%, 37\% and 13\% at $\lambda = 5{\cdot}10^4$, $10^5$ and $3{\cdot}10^5$ (an excess of about 350, 420 and 260 bits per token), and the excess does not shrink with longer training.
Many trained scales are in the range where the estimate is biased. In these three codecs, 76\%, 65\% and 36\% of the components have $\beta_k < 0.2$ (medians 0.11, 0.15 and 0.23), and 43\% to 60\% of the components of \method{}, depending on the entropy model.
Training on the entropy of the rounded code under Eq.~\eqref{eq:laplace} would remove this bias; we have not tried it yet.

\FloatBarrier
\section{Additional Ablations}
\label{app:offline}

\begin{figure*}[!ht]
\centering
\includegraphics[width=\textwidth]{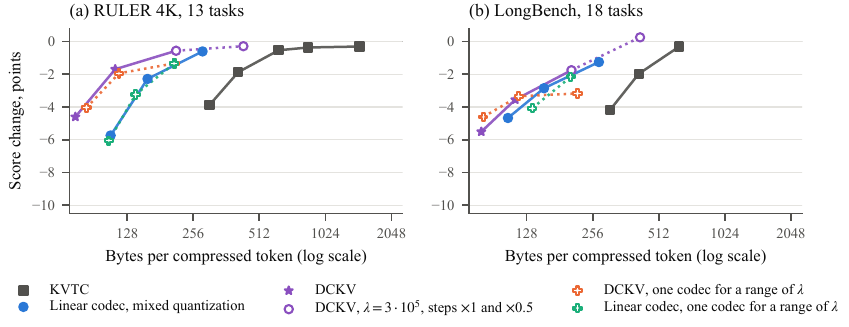}
\caption{As Figure~\ref{fig:rate-quality}, for two ways to change the rate without a codec per rate: \method{} with halved steps, and one codec trained for a range of $\lambda$, with the additions (\method{}) and without them (linear codec).}
\label{fig:variants}
\end{figure*}

\paragraph{Step multiplier instead of retraining.}
Scaling all steps of a trained codec by one factor moves it along the rate axis without training.
For the linear codec at $\lambda = 3{\cdot}10^5$, halving the steps adds 0.51 points on RULER 4K (9/3/1; $p{=}0.02$) at 521 against 282 bytes per token, and quartering them scores 0.40 points lower on RULER 4K than halving them (0/9/4) at 762 bytes.
The linear codec trained at $\lambda = 10^5$ with halved steps scores like the one trained at $3{\cdot}10^5$ ($+0.13$ points; 8/1/4) but needs 24\% more bytes (350 against 282).
Offline, one global factor explains 90--96\% of the log-step difference between the three linear codecs (1.45, 1.66 and 2.41, close to step $\propto \lambda^{-1/2}$); the rest is a per-component pattern of $\pm$11--22\% that gives the components of small scale relatively coarser steps at stronger compression.
One global multiplier cannot reproduce this pattern, which is why retraining at the target $\lambda$ needs fewer bytes; the encoders and decoders of the three codecs differ by only 1.4--2.4\%.
For \method{} at $\lambda = 3{\cdot}10^5$, halved steps give 433 and 421 bytes per token on RULER 4K and LongBench, 31\% and 34\% fewer than KVTC at $16\times$. The loss is 0.28 points on RULER 4K, and on LongBench the score does not differ significantly from the uncompressed model ($+0.26$; 10/0/8). Against KVTC at $16\times$ the difference is $+0.25$ (4/7/2) and $+0.57$ points (10/1/7).
On RULER 32K it also does not differ significantly from the uncompressed model ($+0.24$; 5/6/2) at 441 bytes per token, 19\% fewer than KVTC at $16\times$ ($+0.25$; 4/7/2).
It then has the request memory of KVTC at $24\times$ on RULER 4K and LongBench ($+0.6\%$ and $+0.4\%$) and scores 1.59 points higher on RULER 4K (9/3/1; $p{=}0.02$) and 2.24 higher on LongBench (17/0/1; $p{<}0.001$).
Table~\ref{tab:variants} and Figure~\ref{fig:variants} give these runs and those of the codecs for a range of $\lambda$ below.

\begin{table*}[!ht]
\centering
\footnotesize
\setlength{\tabcolsep}{2.0pt}
\begin{tabular}{l rrr rrrr}
\toprule
 & \multicolumn{3}{c}{Payload, bytes per token} & \multicolumn{4}{c}{Score change vs.\ uncompressed, points} \\
\cmidrule(lr){2-4}\cmidrule(lr){5-8}
Method & RULER 4K & LongBench & RULER 32K & RULER 4K & LongBench & MMLU & RULER 32K \\
 & & & & $n{=}13$ & $n{=}18$ & $n{=}57$ & $n{=}13$ \\
\midrule
Uncompressed & 12,288 & 12,288 & 12,288 & 94.54 & 42.66 & 59.34 & 94.25 \\
\midrule
DCKV, $\lambda{=}3{\cdot}10^{5}$ & 214 & 205 & 221 & -0.56 & -1.75 & -1.10 & -0.45 \\
\quad steps $\times 0.5$ & 433 & 421 & 441 & -0.28 & +0.26 & -1.34 & +0.24 \\
\midrule
DCKV, range, $\lambda{=}3{\cdot}10^{5}$ & 211 & 218 & -- & -1.33 & -3.17 & -- & -- \\
DCKV, range, $\lambda{=}10^{5}$ & 118 & 118 & 130 & -1.94 & -3.34 & -- & -2.08 \\
DCKV, range, $\lambda{=}5{\cdot}10^{4}$ & 83 & 81 & 88 & -4.05 & -4.61 & -- & -5.42 \\
\midrule
Linear codec, $\lambda{=}3{\cdot}10^{5}$ & 282 & 273 & 288 & -0.60 & -1.25 & -0.79 & -0.08 \\
\quad steps $\times 0.5$ & 521 & 511 & 528 & -0.09 & -0.04 & -- & +0.16 \\
\quad + fitted models & 249 & 241 & 253 & -0.60 & -1.25 & -- & -0.08 \\
\bottomrule
\end{tabular}

\caption{Ways to change the rate without a codec per rate, in the form of Table~\ref{tab:main}: all steps of a trained codec halved, conditional models fitted after training, and \method{} trained as one codec for $\lambda \in [5{\cdot}10^4, 3{\cdot}10^5]$. Indented rows change the first row of their block without retraining. --: not run.}
\label{tab:variants}
\end{table*}

\paragraph{The four codecs of the training run of \method{}.}
In the run of \S\ref{sec:results-ablation} ($\lambda = 3{\cdot}10^5$, 1B tokens, four codecs in one loop), the codec without additions stores 245 bytes per token on RULER 4K, the one with the correction 232 ($-5.5\%$), the one with trained conditional models 224 ($-8.7\%$) and \method{} 214 ($-12.8\%$); each has fewer bytes than the first on all 13 tasks, and the scores are 0.96, 0.71, 0.59 and 0.56 points below the uncompressed model.
Conditional models fitted in closed form to the codec without additions give 214 bytes ($-12.5\%$) with its scores, about as many as \method{}, which stores 0.3\% fewer bytes (fewer on 11 of 13 tasks) and scores 0.40 points higher (5/6/2; $p{=}0.45$).
Refitting the trained models to their own transform in closed form gives 3.5\% more bytes than the trained models.
With halved steps \method{} stores 9.9\% fewer bytes than the codec without additions (433 against 481) at the same score ($-0.02$ points).
The closed-form fit to the codec with a correction trained from scratch gives more bytes than no conditional models; we treat this as an unresolved problem of the fit and do not report those two runs.
On the held-out documents the training loss of Eq.~\eqref{eq:loss} is 3.1\% lower with the correction, 10.0\% with the trained models and 12.7\% with both, in all 51 checks of the second half of the run.

\iffullversion\else
\paragraph{One codec for a range of $\lambda$.}
Inspired by variable-rate video codecs~\citep{li2024dcvcfm}, we also trained one codec for a range of rates.
Every training sequence draws its own $\lambda \in [5{\cdot}10^4, 3{\cdot}10^5]$, uniformly in $\ln\lambda$.
The transform and the entropy models are shared by all $\lambda$, and only the quantization steps depend on it, through three learned step vectors that are interpolated over $\ln\lambda$.
We call such a codec a range codec. Table~\ref{tab:range} gives four of them, trained in one run of 1B tokens.
Without additions, the range codec matches the three separate linear codecs of Table~\ref{tab:main} (BD-rate $+0.8\%$ on RULER 4K).
\method{} for the range has a BD-rate of $-31.0\%$ against the range codec without additions, $-22.4\%$ against the same codec with conditional models fitted after training, and $-69.9\%$ against KVTC.
At $\lambda = 10^5$ it is within 0.25 points of \method{} trained for this $\lambda$ alone, which needs 3.5--6.3\% fewer bytes.
At the upper end of the range the additions do not help.
At $\lambda = 3{\cdot}10^5$ on LongBench, \method{} for the range is 1.01 points below the range codec without additions ($p{=}0.03$) and 1.42 points below \method{} trained for this $\lambda$ alone ($p{=}0.03$).
Because we did not find the cause, we train one codec for each rate.
\fi

\begin{table*}[!ht]
\centering
\footnotesize
\setlength{\tabcolsep}{5pt}
\begin{tabular}{l rr rr rr}
\toprule
 & \multicolumn{2}{c}{$\lambda{=}5{\cdot}10^{4}$} & \multicolumn{2}{c}{$\lambda{=}10^{5}$} & \multicolumn{2}{c}{$\lambda{=}3{\cdot}10^{5}$} \\
\cmidrule(lr){2-3}\cmidrule(lr){4-5}\cmidrule(lr){6-7}
Codec & B/tok. & Change & B/tok. & Change & B/tok. & Change \\
\midrule
No additions & 105.7 & -6.06 & 139.9 & -3.23 & 209.5 & -1.34 \\
\quad + models fitted after training & 87.2 & -6.06 & 117.7 & -3.23 & 207.2 & -1.34 \\
Nonlinear correction & 90.4 & -4.70 & 124.6 & -2.35 & 215.2 & -1.30 \\
Trained conditional models & 97.3 & -4.22 & 126.9 & -2.50 & 190.9 & -1.02 \\
Both: DCKV & 83.4 & -4.05 & 117.6 & -1.94 & 210.7 & -1.33 \\
\midrule
Linear codecs, one per $\lambda$ & 107.6 & -5.74 & 158.5 & -2.28 & 282.3 & -0.60 \\
\bottomrule
\end{tabular}

\caption{RULER 4K results ($n{=}13$) of four codecs trained in one run of 1B tokens, each for $\lambda \in [5{\cdot}10^4, 3{\cdot}10^5]$; $\lambda$ is chosen at evaluation. B/tok.: bytes per compressed token; Change: points against the uncompressed model. Models fitted after training do not change the scores. Last row: the three linear codecs of Table~\ref{tab:main}.}
\label{tab:range}
\end{table*}

\paragraph{Conditional models on a development codec.}
Table~\ref{tab:entropy} keeps a codec fixed and changes only its entropy model.
The codec was trained with the noise proxy for 0.1B tokens at $\lambda = 10^5$; its per-task scores are identical in all six runs ($-3.59$ points, $n{=}13$).
Conditioning on the token embedding saves 6.2--6.5\% of the payload, and adding the two decoded neighbours of the checkerboard order saves 9.9--14.3\%, at 5--65~MiB of extra weights depending on the rank.
Longer rANS streams save another 7\%; we could not measure their cost in decoding time reliably on our shared node.
Fitted to the linear codec of Table~\ref{tab:main} (second block), the token model saves 4.3--4.6\%, the token and neighbour models 6.8--11.7\%, and long streams bring the saving to 15.6\%, again with identical scores.

\begin{table}[!ht]
\centering
\footnotesize
\setlength{\tabcolsep}{3.5pt}
\begin{tabular}{l rrrr}
\toprule
Entropy model & Rank & B/tok. & B/tok.\ change & MiB \\
\midrule
\multicolumn{5}{l}{\emph{Development codec}} \\
Static Laplace & -- & 178.1 & -- & 99.1 \\
Token & 2048 & 166.5 & -6.5\% & 131.9 \\
Token & 256 & 167.1 & -6.2\% & 103.9 \\
Neighbours + token & 1024 & 152.6 & -14.3\% & 163.9 \\
Neighbours + token & 256 & 160.5 & -9.9\% & 139.9 \\
\quad + long streams & 1024 & 141.8 & -20.4\% & 163.9 \\
\midrule
\multicolumn{5}{l}{\emph{Linear codec, $\lambda{=}3{\cdot}10^{5}$}} \\
Static Laplace & -- & 282.3 & -- & 99.1 \\
Token & 2048 & 269.2 & -4.6\% & -- \\
Token & 256 & 270.2 & -4.3\% & -- \\
Neighbours + token & 1024 & 249.2 & -11.7\% & -- \\
Neighbours + token & 256 & 263.1 & -6.8\% & -- \\
\quad + long streams & 1024 & 238.3 & -15.6\% & -- \\
\bottomrule
\end{tabular}

\caption{Conditional models fitted in closed form, on RULER 4K: a development codec and the linear codec of Table~\ref{tab:main}; all rows of a block give identical scores. Long streams: rANS streams of 131,072 symbols. MiB: weights (--: not measured).}
\label{tab:entropy}
\end{table}

\iffullversion\else
\paragraph{Other variants.}
Several variants gave little or nothing, and \method{} does not use them.
A nonlinear correction or corrections for 16 groups of tokens, added to a trained linear codec for 0.3B more tokens, saved 1.3--2.7\% of the bits at about the same training loss.
Separate bases of rank 3072 instead of 1024 added 22\% to the payload for $+0.21$ points on RULER 4K.
The following measurements were made offline, on caches of Qwen3.5-2B with the KL divergence of next-token distributions as the quality measure.
Coding each attention layer with its own basis, conditioned on the reconstructed previous layers, cost 13--27\% more bits than the joint basis at equal KL divergence, because the conditioning layers are themselves reconstructed with error.
A mixture of 16 PCA bases, one per group of tokens, used 18--25\% fewer bits than one basis but needs about 1.2~GB of weights.
On Qwen3.5-0.8B, a learned hyperprior~\citep{balle2018hyperprior} saved at most 1.2\% of the payload on top of the token-conditioned model.
\fi

\paragraph{KVTC closer to the original.}
In an offline measurement, a KVTC built as in the original paper, with exact dynamic programming, blocks of 1 to 1024 consecutive components, a data type and a 16-bit offset and scale per block, and DEFLATE, needed 30--53\% fewer bits at equal KL divergence than KVTC with greedy bit allocation.
It was not compared with the KVTC with dynamic-programming allocation that the benchmarks use.

\FloatBarrier
\section{Training and Evaluation Details}
\label{app:details}

\begin{table}[!ht]
\centering
\small
\begin{tabular}{@{}p{0.27\columnwidth}p{0.67\columnwidth}@{}}
\toprule
Language model & Qwen3.5-2B, bf16, frozen \\
Codec size & $d = 6144$, $n = 2048$ \\
Initialization & joint PCA of up to 200K token positions from 48 calibration documents (32 from FineWeb, 16 from OpenR1-Math; 1,008 to 16,218 tokens, 200,192 in total; also used for KVTC); all steps 1.0 \\
Training data & FineWeb-Edu, a 1M-document subset of its 10B-token sample; 4,096 prompt + 256 continuation tokens per sequence; 100 held-out documents \\
Multipliers $\lambda$ & $3{\cdot}10^5$ (\method{} and its three ablations); $10^5$ and $5{\cdot}10^4$ (\method{}), $10^5$ (codec without additions); $[5{\cdot}10^4, 3{\cdot}10^5]$ (range codecs); $5{\cdot}10^4$, $10^5$, $3{\cdot}10^5$ (linear codecs); $3{\cdot}10^4$, $10^5$, $3{\cdot}10^5$ (noise proxy) \\
Additions & correction: one hidden layer of 256 units, GELU, on encoder and decoder, last layer starts at zero (16~MiB); conditional models of rank 2048 (token) and 1024 (context), 32~MiB each, start from their closed-form fit \\
Range codecs & three step vectors, interpolated over $\ln\lambda$; $\lambda$ of a training sequence uniform in $\ln\lambda$ \\
Optimizer & AdamW without weight decay; peak learning rate $10^{-6}$ for $W_e, W_d$, $10^{-3}$ for $\log\Delta$ and the log scales, $10^{-4}$ for the correction and the conditional models; linear warm-up over 10M tokens; cosine decay to zero at 1B tokens; gradient clipping 1.0 per codec \\
Batch & 2 A100 40GB GPUs; runs of \method{}: 1 sequence per GPU, 8,704 tokens and four codecs per step (three in the second loop), 114,890 steps; linear codecs: 2 sequences per GPU \\
Length & 1B tokens for every run except the noise-proxy run: 33.9 hours (\method{} at $3{\cdot}10^5$ and its three ablations), 25.9 hours (\method{} at $10^5$ and $5{\cdot}10^4$, codec without additions at $10^5$), 16.7 hours (three linear codecs), 26.2 hours (four range codecs, Appendix~\ref{app:offline}); the checkpoint of the last step is evaluated; 1.06B tokens (noise; checkpoints at 0.7B evaluated) \\
Entropy coder & symbols within $\pm 127$, 15-bit frequency tables, streams of 16,384 symbols; component active if $b_k \ge 0.05\,\Delta_k$ \\
Active components & \method{}: all 2,048; linear codec: 1,754 of 2,048 at $\lambda = 5{\cdot}10^4$ at the end of training (training logs) \\
Decoding & greedy; non-thinking chat template for generative benchmarks; MMLU scored by option likelihood \\
LongBench & 18 of 21 tasks (without SAMSum, LCC, RepoBench-P); official metrics re-implemented, Chinese text split into characters \\
Answer lengths & RULER: 128 tokens (needle tasks), 120 (CWE), 50 (FWE), 32 (QA), 30 (VT); LongBench: 32--512 per task; GPQA: 1,000; IFEval: 1,280 \\
\bottomrule
\end{tabular}
\caption{Training and evaluation settings.}
\label{tab:details}
\end{table}

\paragraph{Payload and expected code length.}
The code length that a trained entropy model expects, the entropy of the rounded Laplace distribution with the trained steps and scales, is 1,944 bits per token for the linear codec at $\lambda = 3{\cdot}10^5$ and 1,262 bits for \method{}.
The stored payload on the held-out FineWeb-Edu documents is 2,048 and 1,475 bits, 5\% and 17\% more, and on RULER 4K it is 16\% and 35\% more (282 and 214 bytes per token).
The difference combines two effects that we did not measure separately. The codes of a given text differ from what the model expects, more so on RULER than on the training domain, and the coder adds its own cost (the grid of 576 tables for the mean and the scale, escaped codes at 64 bits each, and the end of every rANS stream).

\FloatBarrier
\section{Paired Comparisons}
\label{app:pairs}

Table~\ref{tab:pairs} gives the paired statistics of the comparisons in \S\ref{sec:results}.
$\Delta$ is the paired difference of A minus B in points over the $n$ tasks that both runs share; W/T/L counts the tasks on which A is better, equal and worse; $p$ is the two-sided sign test over the tasks that differ.
With 13 to 18 tasks, a difference of 1--2 points between strong codecs is often not significant, and we did not correct the p-values for multiple comparisons.

\begin{table*}[!ht]
\centering
\footnotesize
\begin{tabular}{l l rrrr rr}
\toprule
Comparison (A vs B) & Benchmark & $\Delta$ & $n$ & W/T/L & $p$ & B/tok.\ A & B/tok.\ B \\
\midrule
\method{} $\lambda{=}3{\cdot}10^5$ vs uncompressed & RULER 4K & -0.56 & 13 & 3/4/6 & 0.508 & 214 & -- \\
 & LongBench & -1.75 & 18 & 0/1/17 & $<$0.001 & 205 & -- \\
 & RULER 32K & -0.45 & 13 & 3/5/5 & 0.727 & 221 & -- \\
 & MMLU & -1.10 & 57 & 15/3/39 & 0.001 & 196 & -- \\
KVTC $16\times$ vs \method{} $\lambda{=}3{\cdot}10^5$ & RULER 4K & +0.03 & 13 & 5/5/3 & 0.727 & 627 & 214 \\
 & LongBench & +1.44 & 18 & 14/0/4 & 0.031 & 633 & 205 \\
 & RULER 32K & +0.43 & 13 & 6/5/2 & 0.289 & 548 & 221 \\
\method{} $\lambda{=}3{\cdot}10^5$ vs KVTC $24\times$ & RULER 4K & +1.31 & 13 & 8/1/4 & 0.388 & 214 & 410 \\
 & LongBench & +0.22 & 18 & 13/1/4 & 0.049 & 205 & 415 \\
 & RULER 32K & +2.00 & 13 & 9/3/1 & 0.021 & 221 & 356 \\
\method{} $\lambda{=}3{\cdot}10^5$ vs KVTC $32\times$ & RULER 4K & +3.33 & 13 & 10/0/3 & 0.092 & 214 & 304 \\
 & LongBench & +2.42 & 18 & 16/0/2 & 0.001 & 205 & 307 \\
 & RULER 32K & +6.49 & 13 & 12/1/0 & $<$0.001 & 221 & 263 \\
\method{} $\lambda{=}10^5$ vs uncompressed & RULER 4K & -1.69 & 13 & 1/2/10 & 0.012 & 113 & -- \\
 & LongBench & -3.55 & 18 & 4/0/14 & 0.031 & 114 & -- \\
 & RULER 32K & -2.15 & 13 & 2/1/10 & 0.039 & 122 & -- \\
 & MMLU & -0.87 & 57 & 15/10/32 & 0.019 & 101 & -- \\
\method{} $\lambda{=}10^5$ vs KVTC $32\times$ & RULER 4K & +2.19 & 13 & 8/2/3 & 0.227 & 113 & 304 \\
 & LongBench & +0.61 & 18 & 14/1/3 & 0.013 & 114 & 307 \\
 & RULER 32K & +4.79 & 13 & 11/1/1 & 0.006 & 122 & 263 \\
\method{} $\lambda{=}10^5$ vs xKV & RULER 4K & +0.96 & 13 & 4/3/6 & 0.754 & 113 & 3,667 \\
 & LongBench & +0.18 & 18 & 11/0/7 & 0.481 & 114 & 3,151 \\
\method{} $\lambda{=}5{\cdot}10^4$ vs uncompressed & RULER 4K & -4.60 & 13 & 1/1/11 & 0.006 & 74 & -- \\
 & LongBench & -5.51 & 18 & 0/0/18 & $<$0.001 & 80 & -- \\
 & RULER 32K & -6.45 & 13 & 0/0/13 & $<$0.001 & 83 & -- \\
 & MMLU & -0.65 & 57 & 18/7/32 & 0.065 & 66 & -- \\
\method{} $\lambda{=}5{\cdot}10^4$ vs KVTC $32\times$ & RULER 4K & -0.72 & 13 & 5/1/7 & 0.774 & 74 & 304 \\
 & LongBench & -1.34 & 18 & 9/1/8 & 1.000 & 80 & 307 \\
 & RULER 32K & +0.49 & 13 & 5/0/8 & 0.581 & 83 & 263 \\
Separate bases, step 2.0 vs KVTC $24\times$ & RULER 4K & +1.15 & 13 & 9/3/1 & 0.021 & 392 & 410 \\
 & LongBench & +0.93 & 18 & 15/0/3 & 0.008 & 389 & 415 \\
Joint vs separate bases, step 4.0 & RULER 4K & +4.30 & 13 & 9/1/3 & 0.146 & 164 & 182 \\
Linear codec $\lambda{=}10^5$ vs untrained joint basis & RULER 4K & +3.03 & 13 & 10/1/2 & 0.039 & 159 & 164 \\
Mixed quantization vs noise proxy, $\lambda{=}10^5$ & RULER 4K & +0.71 & 13 & 7/4/2 & 0.180 & 159 & 160 \\
 & LongBench & +1.66 & 18 & 13/1/4 & 0.049 & 154 & 154 \\
Mixed quantization vs noise proxy, $\lambda{=}3{\cdot}10^5$ & RULER 4K & +0.11 & 13 & 5/5/3 & 0.727 & 282 & 283 \\
 & LongBench & +0.27 & 18 & 11/0/7 & 0.481 & 273 & 274 \\
\method{} vs no additions, same run, $\lambda{=}3{\cdot}10^5$ & RULER 4K & +0.40 & 13 & 5/6/2 & 0.453 & 214 & 245 \\
 & LongBench & -0.02 & 18 & 8/0/10 & 0.815 & 205 & 237 \\
 & RULER 32K & -0.08 & 13 & 2/7/4 & 0.688 & 221 & 251 \\
\method{} vs no additions, same run, $\lambda{=}10^5$ & RULER 4K & +1.13 & 13 & 7/5/1 & 0.070 & 113 & 134 \\
 & LongBench & +0.77 & 18 & 12/0/6 & 0.238 & 114 & 130 \\
\method{} vs fitted models, same run, $\lambda{=}3{\cdot}10^5$ & RULER 4K & +0.40 & 13 & 5/6/2 & 0.453 & 214 & 214 \\
\bottomrule
\end{tabular}

\caption{Paired comparisons. B/tok.: bytes per compressed token of A and B on the benchmark (--: uncompressed). Fitted models: conditional models fitted in closed form after training to the codec without additions of the same run.}
\label{tab:pairs}
\end{table*}

\FloatBarrier
\section{Benchmarks with Short Prompts}
\label{app:short}

\begin{figure*}[!ht]
\centering
\includegraphics[width=\textwidth]{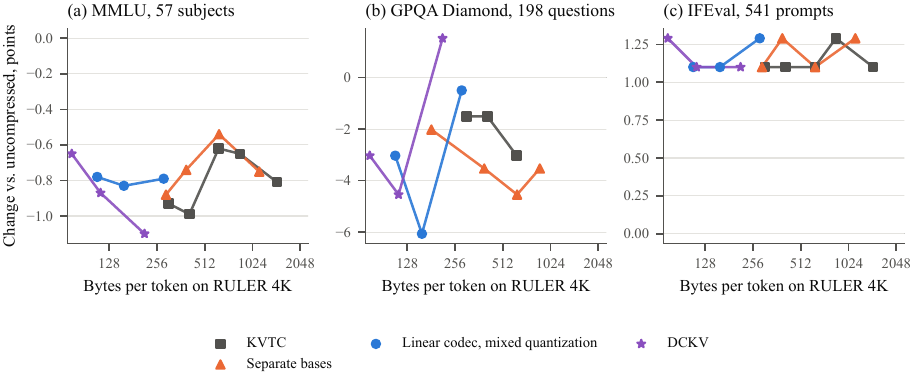}
\caption{Score change on MMLU, GPQA Diamond and IFEval against the payload of the same codec on RULER 4K. Same families and markers as Figure~\ref{fig:rate-quality}. On IFEval every run scores above the uncompressed reference.}
\label{fig:other-benchmarks}
\end{figure*}

Table~\ref{tab:short} gives MMLU and two benchmarks with shorter prompts, and Figure~\ref{fig:other-benchmarks} plots all three against the payload of the same codec on RULER 4K.
GPQA Diamond~\citep{rein2023gpqa} has 198 questions, zero-shot, with a mean prompt of 255 tokens, so one question is 0.5 points.
IFEval~\citep{zhou2023ifeval} has 541 prompts with a mean of 68 tokens, scored by prompt-level strict accuracy.
On GPQA the runs of Table~\ref{tab:short} differ from the uncompressed model by $-6.1$ to $+1.5$ points, with no relation to the payload.

IFEval does not test the codec, since 99.3\% of its prompts fit into the 132 uncompressed positions.
Still, all 29 runs of our codecs and of KVTC score 0.9--1.5 points above the uncompressed reference.
The references are not the cause.
We ran them again on the final evaluation code, and on RULER 4K, LongBench, MMLU, GPQA and IFEval they are identical to the old ones on every task.
The offset comes from the part of our evaluation harness that runs a compressed cache, and we did not find its cause.
On RULER, two near-lossless runs (separate bases of rank 3072 at 1,604 and 3,084 bytes per token) differ from the reference by only $+0.03$ and $-0.01$ points.

\begin{table}[!ht]
\centering
\footnotesize
\setlength{\tabcolsep}{3.5pt}
\begin{tabular}{l rrr}
\toprule
Method & MMLU & GPQA & IFEval \\
\midrule
Uncompressed & 59.34 & 38.38 & 64.70 \\
\midrule
KVTC, $16\times$ & -0.62 & -3.03 & +1.10 \\
KVTC, $24\times$ & -0.99 & -1.51 & +1.10 \\
KVTC, $32\times$ & -0.93 & -1.51 & +1.10 \\
xKV, ranks 192/288 & -0.42 & -2.02 & +1.66 \\
DCKV, $\lambda{=}3{\cdot}10^{5}$ & -1.10 & +1.52 & +1.10 \\
DCKV, $\lambda{=}10^{5}$ & -0.87 & -4.54 & +1.10 \\
DCKV, $\lambda{=}5{\cdot}10^{4}$ & -0.65 & -3.03 & +1.29 \\
DCKV, $\lambda{=}3{\cdot}10^{5}$, steps $\times 0.5$ & -1.34 & +0.00 & +1.47 \\
Linear codec, $\lambda{=}3{\cdot}10^{5}$ & -0.79 & -0.50 & +1.29 \\
Linear codec, $\lambda{=}10^{5}$ & -0.83 & -6.06 & +1.10 \\
Linear codec, $\lambda{=}5{\cdot}10^{4}$ & -0.78 & -3.03 & +1.10 \\
\bottomrule
\end{tabular}

\caption{Score change on the benchmarks with short prompts. The first row gives absolute scores. MMLU: 57 subjects; GPQA: 198 questions; IFEval: 541 prompts.}
\label{tab:short}
\end{table}

\FloatBarrier
\section{Memory and Efficiency}
\label{app:efficiency}

Table~\ref{tab:matched} lists the paired comparisons of \method{} and KVTC at similar request memory that \S\ref{sec:results-matched} discusses.
Table~\ref{tab:efficiency} lists the memory of all runs of Table~\ref{tab:main}, and Table~\ref{tab:timing} the time of three of them.
Peak GPU memory does not fall with compression, because decoding runs on the restored bf16 cache; for \method{} it grows by 182~MiB and for the linear codec by 102~MiB, about the size of the codec weights.
The times come from one timing run of each method, made one after another on one GPU of a shared node.

\begin{table*}[!ht]
\centering
\small
\begin{tabular}{l rrr rrr}
\toprule
 & \multicolumn{3}{c}{RULER 4K (3.7K tokens)} & \multicolumn{3}{c}{RULER 32K (32.2K tokens)} \\
\cmidrule(lr){2-4}\cmidrule(lr){5-7}
Method & Store & TTFT & TPOT & Store & TTFT & TPOT \\
\midrule
Uncompressed & -- & 862 & 45.2 & -- & 4,601 & 42.8 \\
KVTC, $16\times$ & 865 & 121 & 45.7 & 6,333 & 6,849 & 44.4 \\
DCKV, $\lambda{=}3{\cdot}10^{5}$ & 858 & 115 & 44.7 & 4,445 & 287 & 42.1 \\
\bottomrule
\end{tabular}

\caption{Time (ms) of one run of each method, made one after another on one A100 of a shared node; means over about 1,300 prompts. Store: prefill and compression when the prompt is stored. TTFT: from the start of generation to the first token; for a compressed run it includes the decompression of the stored cache, for the uncompressed run the prefill. TPOT: time per output token.}
\label{tab:timing}
\end{table*}

We count codec weights with the model weights, for \method{} and KVTC alike.
If they were charged to a single request instead, \method{}, with 176~MiB of weights, would store less than the uncompressed model only from about 15.4K prompt tokens on; its 156~MiB of extra weights over KVTC at $16\times$ pay off after about 110 stored requests on RULER 4K (1.42~MiB saved per request) and 36 on LongBench (4.37~MiB).

\begin{table*}[!ht]
\centering
\footnotesize
\setlength{\tabcolsep}{3pt}
\begin{tabular}{l rr rr rr}
\toprule
 & \multicolumn{2}{c}{RULER 4K ($n{=}13$)} & \multicolumn{2}{c}{LongBench ($n{=}18$)} & \multicolumn{2}{c}{RULER 32K ($n{=}13$)} \\
\cmidrule(lr){2-3}\cmidrule(lr){4-5}\cmidrule(lr){6-7}
Method & Request, MiB & DCKV $-$ KVTC & Request, MiB & DCKV $-$ KVTC & Request, MiB & DCKV $-$ KVTC \\
\midrule
DCKV, $\lambda{=}3{\cdot}10^{5}$ & 12.1 & & 13.5 & & 18.2 & \\
\midrule
KVTC $16\times$ & 13.5 (+11.7\%)$^{\ast}$ & -0.03 (3/5/5) & 17.8 (+32.4\%)$^{\ast}$ & -1.44 (4/0/14) & 28.1 (+55.0\%)$^{\ast}$ & -0.43 (2/5/6) \\
KVTC $24\times$ & 12.8 (+5.6\%) & +1.31 (8/1/4) & 15.6 (+15.9\%)$^{\ast}$ & +0.22 (13/1/4) & 22.3 (+22.8\%)$^{\ast}$ & +2.00 (9/3/1) \\
KVTC $32\times$ & 12.4 (+2.5\%) & +3.33 (10/0/3) & 14.5 (+7.8\%) & +2.42 (16/0/2) & 19.4 (+7.0\%) & +6.49 (12/1/0) \\
\bottomrule
\end{tabular}

\caption{\method{} ($\lambda{=}3{\cdot}10^5$) against KVTC at similar request memory. Request: request memory and its difference from \method{} ($^{\ast}$ outside $\pm 10\%$, the margin within which we treat request memory as comparable; taken relative to KVTC, the differences mark the same pairs). \method{} $-$ KVTC: paired difference in points, with the tasks where \method{} is higher/equal/lower. Codec weights are not included (Table~\ref{tab:efficiency}).}
\label{tab:matched}
\end{table*}

\begin{table*}[!ht]
\centering
\small
\begin{tabular}{l rrr r rrr}
\toprule
 & \multicolumn{3}{c}{Request memory, MiB} & Weights & \multicolumn{3}{c}{Peak GPU memory, MiB} \\
\cmidrule(lr){2-4}\cmidrule(lr){6-8}
Method & RULER 4K & LongBench & RULER 32K & MiB & RULER 4K & LongBench & RULER 32K \\
\midrule
Uncompressed (bf16) & 54.0 & 137.7 & 387.2 & 0.0 & 4,035 & 4,823 & 7,213 \\
\midrule
KVTC, $16\times$ & 13.5 & 17.8 & 28.1 & 20.1 & 4,057 & 4,846 & 7,234 \\
KVTC, $24\times$ & 12.8 & 15.6 & 22.3 & 13.6 & 4,051 & 4,839 & 7,228 \\
KVTC, $32\times$ & 12.4 & 14.5 & 19.4 & 10.4 & 4,047 & 4,836 & 7,224 \\
xKV, ranks 192/288 & 22.9 & 42.4 & -- & 0.0 & 4,035 & 4,822 & -- \\
\midrule
DCKV, $\lambda{=}3{\cdot}10^{5}$ & 12.1 & 13.5 & 18.2 & 176.1 & 4,217 & 5,006 & 7,394 \\
DCKV, $\lambda{=}10^{5}$ & 11.8 & 12.5 & 15.1 & 176.1 & 4,217 & 5,006 & 7,394 \\
DCKV, $\lambda{=}5{\cdot}10^{4}$ & 11.6 & 12.2 & 13.9 & 176.1 & 4,190 & 4,978 & 7,366 \\
\midrule
Linear codec, $\lambda{=}3{\cdot}10^{5}$ & 12.4 & 14.2 & 20.2 & 99.1 & 4,137 & 4,926 & 7,314 \\
Linear codec, $\lambda{=}10^{5}$ & 11.9 & 13.0 & 16.4 & 99.1 & 4,137 & 4,926 & 7,314 \\
Linear codec, $\lambda{=}5{\cdot}10^{4}$ & 11.8 & 12.5 & 14.8 & 84.8 & 4,124 & 4,912 & 7,300 \\
\bottomrule
\end{tabular}

\caption{Memory of the runs of Table~\ref{tab:main} on one A100 with one request at a time. Request memory (\S\ref{sec:method-memory}) includes the recurrent state (9.84~MiB); for the uncompressed model it also holds the generated tokens (under 1\%), and for xKV the bases that it computes for each prompt. Weights: codec weights loaded with the model; Peak: peak GPU memory. LongBench: 18 tasks. Weights of \method{}: fp32 parameters of the final checkpoint; the other weights are as loaded, with coding tables (99.1~MiB for the linear codec, of which 96.0 are parameters, and less at $\lambda{=}5{\cdot}10^4$, which keeps 1,754 of 2,048 components).}
\label{tab:efficiency}
\end{table*}

\FloatBarrier
\section{Additional Figures}
\label{app:figures}

\begin{figure*}[!ht]
\centering
\includegraphics[width=\textwidth]{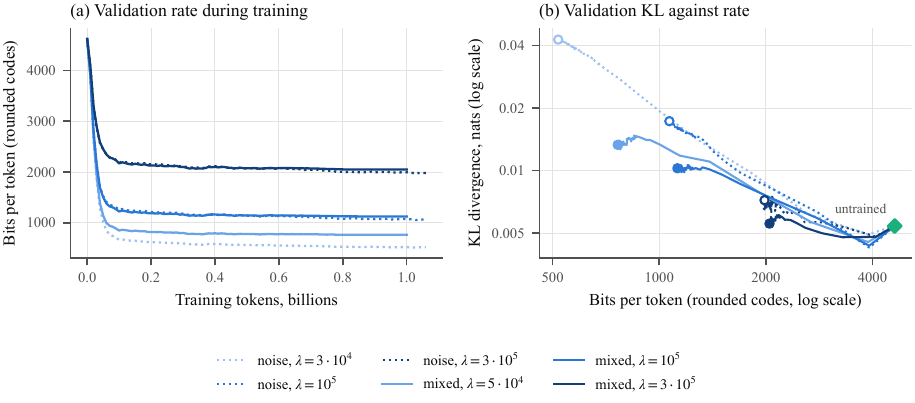}
\caption{Validation during training of the linear codecs, with mixed quantization and with the noise proxy. (a) Rate of rounded codes against training tokens. (b) KL divergence against rate; each curve starts at the untrained codec (diamond), and circles mark the end of training (hollow: noise proxy).}
\label{fig:training-curves}
\end{figure*}

Figure~\ref{fig:training-curves} shows the validation rate and KL divergence of the two training runs of the linear codec, measured every 10M tokens on the 100 held-out documents with rounded codes.
Most of the drop in rate happens within the first 0.1B tokens; after that it falls by a further 6--10\% with mixed quantization and 9--22\% with the noise proxy.
At 1B tokens and the same $\lambda$, the noise-proxy run has a 3--5\% lower rate but a higher KL divergence, 0.0172 against 0.0103 at $\lambda = 10^5$ and 0.0066 against 0.0056 at $\lambda = 3{\cdot}10^5$; at $\lambda = 3{\cdot}10^5$ the mixed run keeps about the KL divergence of the untrained codec (0.0056 against 0.0054) at 44\% of its rate.

\FloatBarrier
\section{Per-Task Results}
\label{app:per-task}

\begin{figure*}[!ht]
\centering
\includegraphics[width=\textwidth]{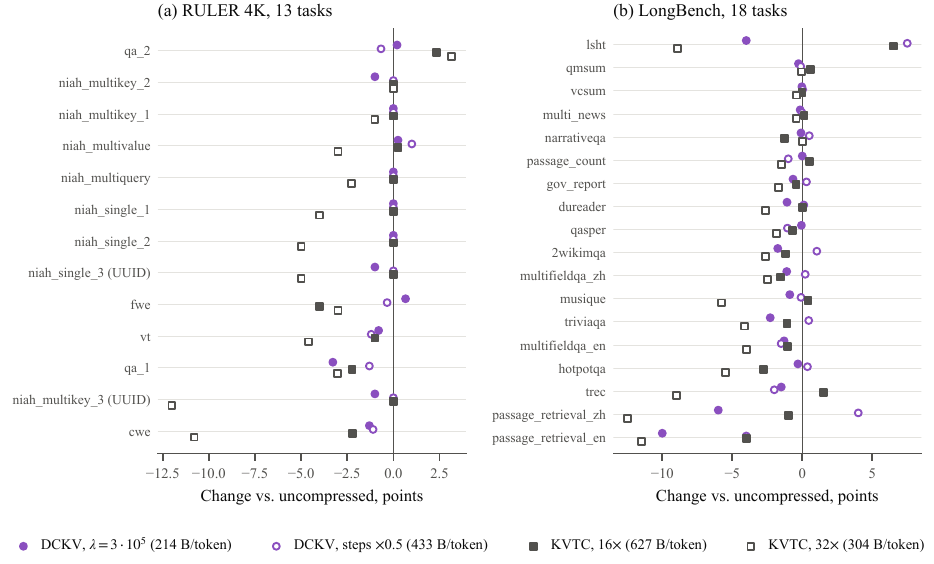}
\caption{Score change per task for \method{} ($\lambda{=}3{\cdot}10^5$) with its trained steps and with halved steps and for KVTC at $16\times$ and $32\times$ (payloads on RULER 4K). Tasks are sorted by the mean change of the four, the hardest at the bottom. UUID: needles with UUID values.}
\label{fig:per-task}
\end{figure*}

Tables~\ref{tab:per-task-ruler} and~\ref{tab:per-task-longbench} list the change of every task against the uncompressed model.

\begin{table*}[!ht]
\centering
\footnotesize
\resizebox{\textwidth}{!}{\begin{tabular}{l r r r r r r r r r}
\toprule
Task & Uncompr. & DCKV & DCKV $\times 0.5$ & Linear $3{\cdot}10^{5}$ & Linear $10^{5}$ & Linear $5{\cdot}10^{4}$ & KVTC $16\times$ & KVTC $24\times$ & KVTC $32\times$ \\
\midrule
cwe & 96.7 & -1.3 & -1.1 & -1.5 & -7.5 & -10.7 & -2.2 & -6.7 & -10.8 \\
fwe & 89.7 & +0.7 & -0.3 & -1.3 & -5.0 & -6.7 & -4.0 & -4.7 & -3.0 \\
niah\_multikey\_1 & 100.0 & +0.0 & +0.0 & -1.0 & -3.0 & -1.0 & +0.0 & +0.0 & -1.0 \\
niah\_multikey\_2 & 100.0 & -1.0 & +0.0 & -1.0 & +0.0 & -1.0 & +0.0 & +0.0 & +0.0 \\
niah\_multikey\_3 & 100.0 & -1.0 & +0.0 & -1.0 & -7.0 & -19.0 & +0.0 & -4.0 & -12.0 \\
niah\_multiquery & 100.0 & +0.0 & +0.0 & +0.0 & -0.7 & -1.5 & +0.0 & -0.7 & -2.2 \\
niah\_multivalue & 99.0 & +0.3 & +1.0 & +0.0 & -1.0 & -2.7 & +0.3 & -1.2 & -3.0 \\
niah\_single\_1 & 100.0 & +0.0 & +0.0 & -1.0 & -1.0 & -1.0 & +0.0 & -2.0 & -4.0 \\
niah\_single\_2 & 100.0 & +0.0 & +0.0 & -1.0 & +0.0 & -3.0 & +0.0 & -2.0 & -5.0 \\
niah\_single\_3 & 100.0 & -1.0 & +0.0 & +0.0 & -3.0 & -13.0 & +0.0 & +0.0 & -5.0 \\
qa\_1 & 82.6 & -3.3 & -1.3 & -0.2 & +1.2 & -3.6 & -2.2 & -2.3 & -3.0 \\
qa\_2 & 62.3 & +0.2 & -0.7 & +1.4 & +1.5 & -1.0 & +2.3 & +3.0 & +3.2 \\
vt & 98.8 & -0.8 & -1.2 & -1.2 & -4.2 & -10.4 & -1.0 & -3.6 & -4.6 \\
\midrule
Mean change, $n{=}13$ & 94.54 & -0.56 & -0.28 & -0.60 & -2.28 & -5.74 & -0.53 & -1.87 & -3.88 \\
\bottomrule
\end{tabular}
}
\caption{RULER 4K per task: score of the uncompressed model and change of each configuration, in points. \method{}: $\lambda{=}3{\cdot}10^5$; $\times 0.5$: halved steps. cwe: common-word extraction; fwe: frequent-word extraction; niah: needle in a haystack; qa: question answering; vt: variable tracking. The last row is the paired mean change over the 13 tasks.}
\label{tab:per-task-ruler}
\end{table*}

\begin{table*}[!ht]
\centering
\footnotesize
\resizebox{\textwidth}{!}{\begin{tabular}{l r r r r r r r r r}
\toprule
Task & Uncompr. & DCKV & DCKV $\times 0.5$ & Linear $3{\cdot}10^{5}$ & Linear $10^{5}$ & Linear $5{\cdot}10^{4}$ & KVTC $16\times$ & KVTC $24\times$ & KVTC $32\times$ \\
\midrule
2wikimqa & 34.1 & -1.7 & +1.0 & +2.1 & -1.3 & -4.2 & -1.2 & -1.0 & -2.6 \\
dureader & 24.4 & -1.1 & +0.1 & -0.4 & -1.5 & -1.6 & +0.0 & -1.4 & -2.6 \\
gov\_report & 30.0 & -0.7 & +0.3 & -0.8 & -0.9 & -1.1 & -0.4 & -0.9 & -1.7 \\
hotpotqa & 46.8 & -0.3 & +0.4 & -0.3 & -2.1 & -6.2 & -2.8 & -5.9 & -5.5 \\
lcc$^{\dagger}$ & 2.0 & +1.8 & +1.5 & +1.5 & +1.4 & +1.2 & +1.0 & +1.9 & +2.3 \\
lsht & 31.0 & -4.0 & +7.5 & -0.5 & -1.0 & -2.0 & +6.5 & +5.0 & -8.9 \\
multi\_news & 24.5 & -0.2 & +0.0 & -0.3 & -0.5 & -0.4 & +0.1 & -0.3 & -0.4 \\
multifieldqa\_en & 53.5 & -1.3 & -1.5 & -1.7 & -2.7 & -2.6 & -1.1 & -2.8 & -4.0 \\
multifieldqa\_zh & 65.3 & -1.1 & +0.2 & +0.5 & -2.0 & -1.3 & -1.6 & -1.5 & -2.5 \\
musique & 30.5 & -0.9 & -0.1 & +0.5 & -4.8 & -8.5 & +0.4 & -1.7 & -5.8 \\
narrativeqa & 22.4 & -0.1 & +0.5 & -0.3 & +0.8 & -0.1 & -1.3 & -0.6 & +0.0 \\
passage\_count & 5.5 & +0.0 & -1.0 & -1.0 & -0.5 & +1.5 & +0.5 & -1.5 & -1.5 \\
passage\_retrieval\_en & 87.0 & -10.0 & -4.0 & -6.5 & -21.5 & -26.0 & -4.0 & -4.5 & -11.5 \\
passage\_retrieval\_zh & 82.0 & -6.0 & +4.0 & -8.0 & -4.5 & -19.5 & -1.0 & -10.5 & -12.5 \\
qasper & 42.4 & -0.1 & -1.1 & -1.4 & -0.5 & -2.3 & -0.7 & -2.6 & -1.8 \\
qmsum & 21.5 & -0.3 & -0.1 & -0.2 & -1.0 & -0.1 & +0.6 & +0.6 & -0.1 \\
repobench-p$^{\dagger}$ & 0.1 & +1.0 & +0.5 & +0.6 & +0.6 & +0.5 & +1.0 & +2.0 & +3.0 \\
samsum$^{\dagger}$ & 33.2 & +3.4 & +2.4 & +2.2 & +3.3 & +3.0 & +3.2 & +2.6 & +2.3 \\
trec & 70.0 & -1.5 & -2.0 & -2.0 & -6.0 & -5.5 & +1.5 & -3.3 & -9.0 \\
triviaqa & 83.2 & -2.3 & +0.5 & -2.5 & -1.4 & -3.8 & -1.1 & -2.3 & -4.1 \\
vcsum & 13.6 & +0.0 & +0.0 & +0.0 & +0.0 & -0.4 & +0.0 & -0.4 & -0.4 \\
\midrule
Mean change, $n{=}18$ & 42.66 & -1.75 & +0.26 & -1.25 & -2.85 & -4.67 & -0.31 & -1.97 & -4.17 \\
\bottomrule
\end{tabular}
}
\caption{LongBench per task: score of the uncompressed model and change of each configuration, in points. $^{\dagger}$Left out of all means: LCC and RepoBench-P score near zero for every configuration in our harness, and on SAMSum and RepoBench-P every run of our codecs and of KVTC scores above the uncompressed reference (\S\ref{sec:setup}). The last row is the paired mean change over the other 18 tasks.}
\label{tab:per-task-longbench}
\end{table*}

\end{document}